\pdfoutput=1

\documentclass[11pt]{article}

\usepackage[]{acl}

\usepackage{times}
\usepackage{latexsym}

\usepackage[T1]{fontenc}

\usepackage[utf8]{inputenc}

\usepackage{microtype}

\usepackage{inconsolata}
\usepackage{hyperref}
\usepackage{url}
\usepackage{amsmath,amsthm,amssymb}
\usepackage{enumitem}
\usepackage{algorithmic}
\usepackage{mathtools}
\usepackage{multirow}
\usepackage{booktabs}
\usepackage{graphicx}
\usepackage{caption}
\usepackage{newfloat}
\usepackage{subcaption}
\usepackage{float}
\usepackage{dsfont}
\usepackage{array}
\usepackage{wrapfig}
\usepackage{tikz}
\usepackage{svg}
\usepackage{catchfilebetweentags}
\usepackage[ruled, lined, linesnumbered, commentsnumbered, longend]{algorithm2e}

\newcolumntype{P}[1]{>{\centering\arraybackslash}p{#1}}

\newcommand{\anli}{{ANLI}}
\newcommand{\fever}{{FEVER}}
\newcommand{\wanli}{{WaNLI}}
\newcommand{\entailer}{{Entailer}}
\newcommand{\ecqa}{{ECQA}}
\newcommand{\race}{{RACE}}
\newcommand{\cosmosqalong}{{Cosmos QA}}
\newcommand{\siqalong}{{SocialIQA}}
\newcommand{\cosmosqa}{{CosQA}}
\newcommand{\siqa}{{SIQA}}
\newcommand{\dream}{{DREAM}}
\newcommand{\boolq}{{BoolQ}}

\newcommand{\arce}{{ARC-e}}
\newcommand{\arcc}{{ARC-c}}

\newcommand{\roberta}{{RoBERTa}}
\newcommand{\entailerm}{{Entailer}}

\newcommand{\entailermx}{{Entailer-11B}}

\newcommand{\flantxxl}{{Flan-T5-xxl}}
\newcommand{\gptthree}{{GPT-3.5}}
\newcommand{\gptfour}{{GPT-4}}

\newcommand{\ourclass}{{Flan-T5-xxl + \textit{Class}}}
\newcommand{\ourrank}{{Flan-T5-xxl + \textit{Rank}}}

\newcommand{\ultwo}{{UL2}}
\newcommand{\codexone}{{Codex-001}}
\newcommand{\lamda}{{LaMDA-137B}}
\newcommand{\chatgpt}{{ChatGPT}}

\DeclareFloatingEnvironment[
fileext=lob,
listname={List of Boxes},
name=Box,
placement=htp,
]{BOX}

\title{Are Machines Better at Complex Reasoning? Unveiling Human-Machine Inference Gaps in Entailment Verification}

\author{ 
\textbf{Soumya Sanyal}\textsuperscript{{$1$}}\quad
\textbf{Tianyi Xiao}\textsuperscript{{$1$}}\quad
\textbf{Jiacheng Liu}\textsuperscript{{$2$}}\\
\textbf{Wenya Wang}\textsuperscript{{$3$}}\quad
\textbf{Xiang Ren}\textsuperscript{{$1$}} \\
{\textsuperscript{$1$}University of Southern California} {\textsuperscript{$2$}University of Washington}\\ {\textsuperscript{$3$}Nanyang Technological University, Singapore} \\
{\texttt{soumyasa@usc.edu}}
}

\begin{document}
\maketitle

\begin{abstract}
Making inferences in text comprehension to understand the meaning is essential in language processing.
This work studies the entailment verification (EV) problem of multi-sentence premises that requires a system to make multiple inferences implicitly.
Studying EV for such complex premises is important because modern NLP problems, such as detecting \textit{inconsistent} model-generated rationales, require complex multi-hop reasoning.
However, current textual inference datasets mostly contain short premises that only partially focus on these challenges.
To address this, we compile an EV benchmark that includes datasets from three NLP domains (NLI, contextual QA, and rationales) containing multi-sentence premises.
On benchmarking humans and LLMs, we find that LLMs are better than humans in multi-hop reasoning across extended contexts, while humans perform better in simple deductive reasoning tasks.
We also finetune a Flan-T5 model\footnote{\url{https://huggingface.co/soumyasanyal/entailment-verifier-xxl}} for EV using two training objectives to obtain a strong open-source model that outperforms \gptthree{} and rivals \gptfour{}.
Finally, we use this model to filter out inconsistent model-generated rationales in self-consistency decoding, resulting in a 6\% accuracy improvement on average across three MCQ datasets.
\end{abstract}
\ExecuteMetaData[sections/figures.tex]{motivation}

\ExecuteMetaData[sections/tables.tex]{datasetComparisons}

\section{Introduction}
A prevailing notion in cognitive psychology exists that humans make numerous inferences to understand discourse and text \cite{GARNHAM1989153}.
These inferences play a crucial role in linking information from disparate sections of a text to establish its literal meaning and are closely associated with \textit{reasoning}.
With the recent applications of Large Language Models (LLMs) \cite{devlin2018bert,raffel2019exploring,gpt3,flant5} in NLP tasks that require inference skills \cite{survey_text_classification}, it is thus essential to understand and improve upon the limitations of LLMs concerning different aspects of language inferences.

In this work, we focus on the task of \textit{entailment verification} (EV) that classifies whether a given context supports a hypothesis.
To assert the validity of a hypothesis, a system has to make multiple inferences from the given context and its internal knowledge, which requires complex multi-hop reasoning.\footnote{In a multi-hop reasoning instance, a system has to infer implicit inferences by combining information from the premise, to predict the support of a hypothesis.}
Verifying the entailment of such complex premises has important applications in rationale-generating LLMs, such as chain-of-thoughts (CoT) \cite{wei2022chain}, that typically generate multi-sentence rationales with incomplete information and suffer from inconsistent reasoning \cite{ye2022the}.
While EV is similar to Natural Language Inference (NLI) \cite{rte_paper,Manning2009NaturalLI}, some interesting challenges distinguish it from NLI.
Many existing textual inference datasets, such as SNLI \cite{snli}, MNLI \cite{mnli}, etc., mostly contain short sentence premises that only partially encapsulate the challenges of multi-sentence premises requiring complex reasoning.
Predicting the entailment of such complex premise-hypothesis pairs often requires multi-hop reasoning, inferring missing information, etc., which is lacking from standard NLI datasets \cite{gururangan-etal-2018-annotation,mccoy-etal-2019-right}.
Thus, studying EV in the context of modern LLMs is essential and is currently missing.

To this end, we first compile an evaluation benchmark to study the EV problem by selecting multiple datasets across three categories: NLI, contextual QA, and rationales.
As shown in Table \ref{tab:dataset_comparisons}, the datasets typically contain multi-sentence premises that require inferring different types of knowledge to predict the entailment.
Thus, this benchmark is more complex than standard NLI datasets and can be used as a new evaluation benchmark.

Next, we evaluate LLMs and humans on this benchmark and make some interesting observations.
Cognitive studies \cite{doi:10.1073/pnas.1104666108,Cowan_2001} have shown that an average human brain has a limited capacity to retain only four chunks in short-term memory, indicating a limitation of human inference abilities over long contexts.
Our analysis shows that LLMs are indeed stronger than humans at tasks that involve multi-hop reasoning across long contexts, reinforcing this human limitation.
In contrast, humans are better in cases that require simple deductive reasoning using substitutions, negations, etc., indicating that current LLMs lack consistency along these reasoning aspects.
Further, humans and LLMs perform comparably in instances requiring inferring missing knowledge.
These findings are depicted in Figure \ref{fig:motivation} with motivating examples.

Additionally, on comparison between LLMs, we find that models finetuned on a specific dataset category are usually strong within the category but don't generalize well on unseen categories.
In contrast, instruction-finetuned models are better on average, with the best performing model, \gptfour{}, achieving 0.79 macro-F1 and outperforming the best-open-sourced model \flantxxl{} by 0.08 macro-F1 on average.
In order to bridge this gap, we finetune a Flan-T5 \cite{flant5} model on a training subset containing datasets from each category of the above data collection.
To this end, we explore two different training approaches: a \textit{classification-based} finetuning that learns to directly predict the label and a \textit{ranking-based} finetuning that learns to rank the most supported hypothesis from a given pair of hypotheses for a given premise.
Ranking-based finetuning is better than classification, specifically in contextual QA datasets, as it can learn a softer decision boundary.
Overall, our fine-tuned Flan-T5 outperforms \gptthree{}, baseline Flan-T5, and performs comparably to \gptfour{} on the benchmark, thus providing a strong open-sourced model for entailment verification.

Finally, we demonstrate the utility of our finetuned models on a downstream application of filtering unfaithful model-generated explanations.
Self-consistency (SC) \cite{wang2023selfconsistency} decoding samples multiple model-generated reasoning paths from the LLM decoder and aggregates them to predict the most consistent answer.
We use our finetuned Flan-T5 to filter out non-entailed reasoning chains before aggregating the final prediction, which leads to 6\% performance improvement on average across three MCQ datasets.

Overall, our contributions can be summarized as follows:
\begin{enumerate}[itemsep=1pt,topsep=1pt,leftmargin=*]
    \item Using existing resources, we develop a benchmark for Entailment Verification (EV) of multi-sentence premises for both training and evaluation purposes.
    \item We compare and contrast humans and \gptfour{} on this benchmark and conclude that humans are more robust on simple deductive reasoning while LLMs excel at complex inferences.
    \item We use a ranking-based objective to finetune \flantxxl{} model for the EV task, achieving comparable performance to \gptfour{} and demonstrate the utility of our model in filtering non-entailed CoT rationales.
\end{enumerate}
\section{Entailment Verification using LLMs}
\label{sec:human_alignment}
In this section, we formally define the task of entailment verification (EV), the datasets used to create the evaluation benchmark, and the evaluation procedure for evaluating different LLM baselines.

For a given premise $p$ and a hypothesis (or claim) $h$, the task of entailment verification is to determine whether the context has information that directly confirms the hypothesis or not, i.e., whether the hypothesis follows from the information present in the context. This is a binary classification task defined as $f(p, h) = \{\textrm{\textit{support}}, \textrm{\textit{not support}}\}$, where $f$ is a classifier (human/LLM).

\subsection{Evaluation Benchmark}
\label{sec:datasets}
In this section, we list some desirable properties we want to include in the EV benchmark, followed by details of the dataset categories we select.
\begin{itemize}[itemsep=1pt,topsep=1pt,leftmargin=*]
	\item \textbf{Type of Premise}: Typically, NLI datasets, such as SNLI, MNLI, etc., do not contain more than one sentence in the premise, potentially leading to shortcut learning. In contrast, we focus more on multi-sentence premises that require complex reasoning. We also consider datasets where the premise is a \textit{rationale}, i.e., the premise is not just a logical precursor to the hypothesis but rather an explanation. This tests the ability to evaluate model-generated rationales \cite{wei2022chain}.
	\item \textbf{Type of Knowledge}: Often, one or more information in the premise needs to be used to predict support. We categorize this information as entity-grounded, commonsense, or localized. Entity-grounded knowledge consists of information about entities and other general knowledge verifiable on the internet. These can be facts about general science, history, etc., or details of some known person, event, etc. It is possible to infer this information even if not mentioned in the premise. The commonsense knowledge is typically all information about everyday life that humans use implicitly but cannot always be verified online. This information is often missing from the premise and has to be inferred implicitly. Lastly, localized information is all other knowledge provided for understanding the events, people, or items mentioned in the premise that are not grounded to any known entity. This information depends on the premise's specific context and, thus, is impossible to infer unless stated explicitly. Please refer to Table \ref{tab:knowledge_examples} for examples of each knowledge type.
\end{itemize}

\ExecuteMetaData[sections/tables.tex]{knowledgeExamples}

\noindent We consider three data sources for creating the entailment verification benchmark, amounting to 10 datasets in total. In Table \ref{tab:dataset_comparisons}, we compare these datasets across the desirable characteristics mentioned earlier. Please refer to Appendix \ref{app:dataset_statistics} for more details on the datasets used.

\paragraph{Natural Language Inference} Given the close connection between NLI and EV, it is an obvious choice to consider appropriate NLI datasets for the benchmark. To convert an NLI dataset for EV, we merge the \textit{neutral} and \textit{contradict} labels to the \textit{not support} label. We use the following NLI datasets in our benchmark: \wanli{} \cite{wanli}, \fever{} \cite{fever_nli}, and \anli{} \cite{anli}.

\ExecuteMetaData[sections/tables.tex]{humanResults}

\paragraph{Contextual QA} Next, we consider multiple-choice question-answering datasets where the task is to answer a question based on a given context and some options. We use a QA-to-statement converter model \cite{chen-etal-2021-nli-models} to generate a hypothesis statement for each question option pair. Then, the hypothesis corresponding to the correct choice is marked as ``\textit{support}'', while the rest are marked as ``\textit{not support}''. This process is depicted in Figure \ref{fig:contextual_qa} in Appendix \ref{app:dataset_statistics}. We include the following datasets from this category: \cosmosqalong{} (\cosmosqa{}) \cite{cosmosqa}, \siqalong{} (\siqa{}) \cite{socialiqa}, \dream{} \cite{dream}, \boolq{} \cite{boolq}, and \race{} \cite{race}.

\paragraph{Rationale} Lastly, we consider data sources where human-annotated explanations are available that justify the original hypothesis. In this case, we use the rationales as the premise. We use the following datasets: \entailer{} \cite{entailer}, and \ecqa{} \cite{ecqa}.

\subsection{Evaluation Metric}
\label{sec:eval_metric}
We use the macro-F1 score as the primary evaluation metric for comparing LLMs on the entailment verification task because there are label imbalances in our evaluation datasets. The macro-F1 score computes the unweighted mean of F1 scores for each class, ensuring equal importance for each class irrespective of the label statistics. Please refer to Appendix \ref{app:majority_baseline} for more discussions on the label imbalance of each dataset.

\subsection{LLM Evaluation Setup}
\label{sec:baselines}
We evaluate two types of LLMs on the task of EV, as categorized below:
\paragraph{Task-finetuned LLMs} In this, the models considered are already finetuned on some subset of the benchmark. We evaluate two models: \roberta{} \cite{roberta} (finetuned on NLI datasets) and \entailermx{} \cite{entailer} (finetuned on \entailer{} dataset). Refer to Appendix \ref{app:task_finetuned} for more details on the evaluation setup for these models.

\begin{BOX}
	\centering
	\noindent\fbox{%
		\parbox{0.95\columnwidth}{%
			\textbf{Premise}: \{\textit{premise}\}\\
			\textbf{Hypothesis}: \{\textit{hypothesis}\}\\
			\textbf{Question}: Given the premise, is the hypothesis correct?\\
			\textbf{Answer}:
		}%
	}
	\caption{\label{box:prompt} Prompt used to evaluate instruction-finetuned LLMs for entailment verification.}
	\vspace{-0.4cm}
\end{BOX}
\paragraph{Instruction-finetuned LLMs} These are language models trained on a collection of NLP tasks described using instructions, leading to generalization abilities to solve unseen tasks described using new instructions. Here, we evaluate \flantxxl{} \cite{flant5}, \gptthree{} \cite{gpt3}, and \gptfour{} \cite{gpt4} models. To compute the label, we first modify a given premise-hypothesis pair $(p, h)$ into a prompted input $\mathcal{P}$ using the prompt template as shown in Box \ref{box:prompt}. Next, we compute a score $s$ as defined below:
\begin{equation}
\label{eq:score}
s(p,h) = \frac{p_{LLM}({\textrm{``Yes''} | \mathcal{P}})}{p_{LLM}({\textrm{``Yes''} | \mathcal{P}}) + p_{LLM}({\textrm{``No''} | \mathcal{P}})},
\end{equation}
where $p_{LLM}(\cdot | \mathcal{P})$ is the model's probability distribution over the vocabulary. If the score $s$ is higher than a threshold (typically set to $0.5$ in all our experiments), we assign the label \textit{support}, else we assign the label \textit{not support}. For \gptfour{} evaluation, we directly check for the ``Yes'' / ``No'' label prediction as the token probabilities are not accessible via the API. Please refer to Appendix \ref{app:instruction_finetuned} for more details about the models and ablations on prompts.

\section{Evaluation of Humans and LLMs}
\label{sec:human_performance}
First, we randomly sample 100 instances from each dataset (i.e., 1000 instances in total) and conduct a human evaluation on this subset to estimate average human performance. Please refer to Appendix \ref{app:human_ann_eval} for more details on the annotation procedure. Additionally, we evaluate the above LLMs on this sampled subset and report those numbers for fair comparisons with humans. Table \ref{tab:human_results} shows the overall evaluation results.

\subsection{Comparison among LLMs}
\label{sec:res_llms}
We observe that the task-finetuned models (rows 1-2 in Table \ref{tab:human_results}) are weaker on average compared to the instruction-finetuned models (rows 3-5 in Table \ref{tab:human_results}). However, there are some interesting exceptions. \roberta{}, which is finetuned on NLI datasets, performs at par with \gptfour{} on \fever{} and outperforms it on \wanli{} but falls behind on contextual QA and rationale datasets. On the other hand, \entailermx{}, which is finetuned on the \entailer{} dataset, outperforms \gptfour{} on \entailer{} but lags on most of the other datasets. This demonstrates that finetuning an LLM using datasets from one of these categories is ineffective in outperforming models trained on more general data. Overall, we observe that \gptfour{} is the best-performing entailment verification model on average and \flantxxl{} is the best open-source LLM for this task.

\subsection{Comparison between Humans and LLMs}
\label{sec:res_humans_llm}
In Table \ref{tab:human_results}, we report the human performance and the corresponding annotation agreement\footnote{We use a pairwise agreement ratio that computes the fraction of matched annotations over all pairs of annotations.} (in parenthesis) for each dataset. We note that the agreement ratios are lowest for Contextual QA datasets, especially \cosmosqa{} and \siqa{}. Questions in these datasets are often based on commonsense scenarios, and sometimes, the support for the wrong hypothesis can be debatable. Please refer to Appendix \ref{app:debatable_example} for examples. Between humans and LLMs, we find that humans beat all the baseline LLMs, except \gptfour{}. This shows that existing open-sourced LLMs are subpar with humans on this task. Additionally, although humans and \gptfour{} perform comparably on average, we note that large misalignments exist in different individual datasets. Specifically, we observe that \anli{}, \cosmosqa{}, \siqa{}, and \entailer{} are the four datasets with $>0.1$ absolute macro-F1 difference. We analyze these four datasets in Section \ref{sec:reasoning_type}.

\ExecuteMetaData[sections/figures.tex]{reasoningAnalysis2}

\ExecuteMetaData[sections/tables.tex]{finetuneResults}

\subsection{Effect of Reasoning Type}
\label{sec:reasoning_type}
We design an analysis to understand further the misalignments between \gptfour{} and humans. First, we categorize the type of reasoning required to predict an entailment into the following four categories:
\begin{itemize}[itemsep=1pt,topsep=1pt,leftmargin=*]
	\item \textbf{Simple Deductive (R1)}: The premise contains sentences that can be minimally combined in one step to predict the support for the hypothesis. This typically tests skills such as substitution, understanding negations, word meanings, etc.
	\item \textbf{Complex Deductive (R2)}: More than one step of reasoning is required to solve the task. Typically, this tests skills like mathematical reasoning, combining multiple information in context, etc.
	\item \textbf{Missing Entity-grounded/Commonsense Knowledge (R3)}: In this, some essential commonsense or entity-grounded knowledge is missing in the premise. Both humans and LLMs can implicitly invoke such information from the memory or the parametric knowledge obtained from pertaining, respectively.
	\item \textbf{Missing Localized Knowledge (R4)}: In this, information very specific to the premise is missing. Typically, this is information about the subjects in the context and is not grounded in any known entities. It is practically impossible for humans or LLMs to infer such information.
\end{itemize}
We note that these categories are mutually exclusive.\footnote{Deductive reasoning implies that the premise has all the necessary information. Thus, any missing knowledge instance falls under inductive reasoning.} Figure \ref{fig:reasoning_analysis2} depicts the aggregated results. The top plot shows the percentage of each reasoning type among 400 samples from \anli{}, \cosmosqa{}, \siqa{}, and \entailer{}, and the bottom plot compares the human and \gptfour{} macro-F1 scores. Please refer to Appendices \ref{app:human_ann_reasoning}, \ref{app:reasoning_examples}, and \ref{app:reasoning_analysis} for details on the annotation setup, examples of each reasoning type, and detailed analysis, respectively.

The first type in Figure \ref{fig:reasoning_analysis2} is simple deductive reasoning ($\sim$ 25\% data). Here, humans perform better than \gptfour{} by a small margin. Instances that require simple deductive reasoning usually use substitutions, negations, paraphrasing, etc., to prove entailment. We find that humans are more robust than \gptfour{} in performing such simple deductive reasoning tasks, which is also observed in prior works \cite{robustlr,nguyen2023negation}.

Next, we find that \gptfour{} significantly outperforms the humans on complex reasoning that constitutes $\sim$ 20\% data. This usually requires two skills: understanding multiple relevant information in the premise and combining them for reasoning. \gptfour{} is likely a stronger context processor, especially for long premises, since it has been trained on long-context data sources \cite{gpt4}. In contrast, cognitive studies \cite{doi:10.1073/pnas.1104666108,Cowan_2001} have shown that an average human brain can retain only four chunks in short-term memory, thus limiting the long-context processing abilities of humans.

Lastly, we observe that approximately 30\% of the data has some missing entity-grounded or commonsense information while $\sim$ 25\% of the data has missing localized information. To correctly predict entailment in such cases, a system needs to infer some missing grounded knowledge while not hallucinating specific localized information not mentioned in the premise. We find that both humans and models are comparable across the reasoning types \textbf{R3} and \textbf{R4}, with \textbf{R4} being more challenging. This shows that both models and humans tend to hallucinate missing localized information.

\section{Training LLMs for Entailment Verification}
\label{sec:training_llm}
In Section \ref{sec:res_humans_llm}, we observed that open-sourced LLMs lack performance compared to humans and close-sourced models such as \gptfour{}. From Section \ref{sec:res_llms}, we know that task-finetuned models perform well on a category when finetuned on data from the same category. Using this insight, we finetune the \flantxxl{} model on the train splits of datasets from each category, resulting in using \anli{} \cite{anli}, \race{} \cite{race}, and \ecqa{} \cite{ecqa} for finetuning. Please refer to Appendix \ref{app:training_data_selection} for more details on our training dataset selection criteria. Next, we describe the finetuning approaches and our key findings.

\subsection{Finetuning Formulations}
This section describes the two finetuning formulations explored in this work.

\paragraph{Classification} This is the standard training paradigm where we finetune a \flantxxl{} model using the training data. We follow the same steps as the evaluation setup to create a prompted input using the prompt format in Box \ref{box:prompt} and then define the cross-entropy loss over the ``Yes'' and ``No'' token logits. We refer to this finetuned model as ``\ourclass{}''.

\ExecuteMetaData[sections/figures.tex]{cotFlowchart}

\paragraph{Ranking}
In this approach, for a given premise and hypothesis pair $(p, h)$, we define a \textit{weaker hypothesis} $h^\prime$ as a statement such that the premise $p$ supports $h$ more strongly than $h^\prime$. Then, for a given triplet $(p, h, h^\prime)$, we formulate the ranking task as predicting the hypothesis that is \textit{more} supported by the premise. Given the triplet $(p, h, h^\prime)$, we define the margin ranking loss as follows:
\begin{equation}
\label{eq:ranking_loss}
\mathcal{L}_{ranking} = \max\{0, s(p, h) - s(p, h^\prime) + m\},
\end{equation}
where $s(p, h)$ is the entailment score as defined in Equation \ref{eq:score}. The key advantage of this formulation over the classification is that ranking, by design, learns a softer decision boundary between the two labels. This can lead to better generalization, especially for contextual QA datasets. Sometimes, the wrong choice can be relatively less favorable compared to the best option in QA instead of being incorrect. As discussed in Section \ref{sec:res_humans_llm}, this is indicated by the low agreement between human raters. Training to hard-classify the hypothesis for such options can be avoided by ranking them with the best hypothesis (corresponding to the right choice), thus learning a softer classification boundary. We refer to the finetuned model using the ranking objective as ``\ourrank{}''. Please refer to Appendix \ref{app:ranking} for more details on the training data collection process for ranking.

\subsection{Findings}
\label{sec:finetuning_results}
Table \ref{tab:finetune_results} shows the evaluation results on the complete evaluation set (i.e., we use all the data points instead of 100 samples per dataset, which was used in Table \ref{tab:human_results}). For our models, we separately average the results for the datasets already seen in training (namely, \anli{}, \race{}, and \ecqa{}) and unseen during training into two columns, seen and unseen, respectively. First, we observe that all our finetuned models are consistently better across nine of ten datasets than the baseline \flantxxl{}. Finetuning improves 0.07 macro-F1 on average over \flantxxl{}. This shows that finetuning is overall beneficial in training the model on the task of entailment verification.

Next, we observe that compared to classification, the ranking formulation is beneficial for the contextual QA datasets \cosmosqa{}, \siqa{}, and \dream{}. This demonstrates that the ranking objective improves contextual QA datasets' generalization, which is expected. Additionally, our ranking model outperforms \gptthree{} and performs comparably to \gptfour{}, with stronger performance on contextual QA datasets and weaker performance on NLI datasets. Thus, \ourrank{} is a strong \textit{open-sourced} model for entailment verification and can be used as an alternative to \gptfour{}.

\ExecuteMetaData[sections/figures.tex]{cot}

\section{Application: Filtering CoT Rationales}
\label{sec:filtering_cot}
Recently, \citet{wang2023selfconsistency} proposed self-consistency (SC), a decoding technique to improve over chain-of-though (CoT) reasoning \cite{wei2022chain} in LLMs, whereby multiple CoT rationales are sampled for a given input instance and a majority voting overall predicted labels is considered as the final prediction. However, generative LLMs can potentially output rationales that are inconsistent \cite{ye2022the}, i.e., the rationale does not support the corresponding model prediction. Such inconsistency can, in turn, degrade the overall self-consistency results. Evaluating the consistency between an explanation and the corresponding prediction can be framed as an \textit{entailment verification} (EV) task, as described below. Please refer to Appendix \ref{app:cot_examples} for examples of consistent and inconsistent CoTs.

\paragraph{Approach} As shown in Figure \ref{fig:cot_application}, we can use a verifier as an intermediate filtering step to filter out the inconsistent rationales before computing the majority vote. For this, we define the generated CoT rationale as the premise and use the QA-to-statement model \cite{chen-etal-2021-nli-models} as defined in Section \ref{sec:datasets} to convert the question and model's prediction into a hypothesis. Next, we calculate the entailment score of all the premise-hypothesis pairs using a verifier (Equation \ref{eq:score}). Finally, we select the top-$k$ rationales for majority voting, discarding the rest. We set $k = 5$ for all our experiments.

\paragraph{Findings} In figure \ref{fig:cot_comparison}, we compare the vanilla SC with the filtering+SC approach described above. Following \cite{wang2023selfconsistency}, we compute the average performance of these methods across three MCQ datasets for four different base CoT models (\ultwo{} \cite{ul2}, \codexone{} \cite{gpt3}, \lamda{} \cite{lamda}, and \chatgpt{}\footnote{ChatGPT refers to the \texttt{gpt-3.5-turbo-0613} model.} \cite{chatgpt}). Please refer to Appendix \ref{app:cot_filtering} for details on the datasets and more comparisons with \flantxxl{}. We observe that filtering leads to a consistent performance gain over SC across all CoT base models. This demonstrates the advantage of the filtering approach. Next, we find that the improvements are more prominent for weaker base models such as \ultwo{} than the stronger ones (\chatgpt{}). For instance, filtering \ultwo{} generated-rationales can even achieve comparable performance with vanilla SC over \lamda{}. In comparison, the gains for filtering \chatgpt{} CoTs are $\sim$ 1.7 \%. This shows that weaker models are prone to generating inconsistent CoTs and thus benefit more from this approach. But at the same time, even stronger models such as \chatgpt{} can still benefit from consistency checks. Please refer to Appendix \ref{app:cot_examples} for examples of filtered CoT rationales.

\section{Related Works}

\paragraph{Natural Language Inference} NLI \cite{rte_paper,Manning2009NaturalLI} is one of the core NLP problems in which the relationship between a premise and hypothesis is classified as either entailment, contradiction, or neutral.
Prior works have mainly trained LLMs and evaluated them on standard NLI datasets \cite{snli,mnli,wang2019glue,anli,fever_nli,wanli}.
Another line of work \cite{mishra-etal-2020-reading,chen-etal-2021-nli-models} has used question-answer-to-NLI conversion \cite{demszky2018transforming} to transform QA datasets into NLI format and solve them.
In fact verification literature \cite{fact_verification_survey,fever}, retrieved pieces of evidence have been used to verify the claim using an entailment verifier \cite{fever_nli,guan2023language}.
Recently, NLI models have been used to verify the entailment of model-generated explanations \cite{entailer,jung-etal-2022-maieutic,mitchell-etal-2022-enhancing}.
In this work, we curate a diverse NLI benchmark for evaluating LLMs and humans by using datasets from all the above NLI applications.

\paragraph{Reasoning in LLMs}
With the advent of general-purpose LLMs \cite{gpt3,flant5,gpt4}, many prompting strategies have been proposed to generate a natural language reasoning along with the model's prediction \cite{wei2022chain,zhou2023leasttomost,yao2023tree,huang-chang-2023-towards}.
Recently, \citet{ye2022the} have found that such generations can sometimes be unreliable due to non-factual and inconsistent reasoning, while \citet{huang2023large} have argued that LLMs struggle to self-correct such issues without external feedback.

Prior works have addressed this limitation by oversampling reasoning chains and marginalizing \cite{wang2023selfconsistency}, using the LLMs itself to recheck their reasoning \cite{madaan2023selfrefine,miao2023selfcheck}, leveraging external knowledge source to verify factuality \cite{zhao-etal-2023-verify}, using deterministic solvers to improve faithfulness \cite{lyu2023faithful}, decomposing the reasoning steps into smaller steps \cite{ling2023deductive}, etc. While the progress is impressive, some of these are either specialized approaches for math-specific datasets or heavily rely on close-sourced LLMs (\gptthree{}, \gptfour{}, etc.) for verification. In contrast, here we focus on natural language datasets and develop an open-sourced model for the verification task.

\paragraph{Ranking Objective}
Prior works have explored the benefits of ranking objective both from a theoretical perspective \cite{NIPS2013_05311655} and specifically for NLP classification tasks \cite{li-etal-2019-learning,briakou-carpuat-2020-detecting}. Our ranking-based objective is inspired by these works that find that margin-based loss can be beneficial in training plausibility estimation models.
\section{Conclusion}
We studied the EV problem in the context of LLMs. Specifically, we sourced datasets across three different categories (NLI, contextual QA, and rationales) and analyzed human and LLM performance on the benchmark. We found that LLMs are better than humans in complex reasoning, while humans are more consistent on simpler reasoning tasks. We also explored different training objectives to finetune open-sourced LLMs for EV. Our finetuned model outperforms \gptthree{}, baseline \flantxxl{}, and is comparable to \gptfour{}. Finally, we applied the EV model to filter out inconsistent model-generated CoTs in self-consistency decoding, achieving improvements over the baseline self-consistency approach.

\section*{Limitations}
Even though our work demonstrates exciting results on entailment verification tasks by finetuning LLMs, several limitations can be potentially improved.

\paragraph{Data Processing} Our strategy to convert a QA pair into a statement using the QA-to-statement converter model can have errors that can cascade both in the evaluation dataset and our fine-tuned models. Better strategies for this pipeline would help with data quality.

\paragraph{Finetuning} We only tried encoder-decoder-based models for finetuning. However, other models with different architectures (like decoder-only) can also be considered in the future.
For computing the entailment score in Equation \ref{eq:score}, we only considered the probability of ``Yes'' and ``No'' tokens in the entire vocabulary. Other alternative expressions like ``YES''/``NO'', ``True''/``False'', etc., can also be considered to make the score more robust.
Finally, our training objective outputs entailment scores instead of directly generating answers. Answer generation as a training objective can be more robust since it is a stricter objective than our scoring technique.

\paragraph{Potential Risks} We using existing datasets, with some post-processing, to train our models. Thus, any issues in the existing dataset in terms of bias, toxicity, etc., can potentially affect our model. Also, since we use a pretrained checkpoint, we also inherit any existing biases in the baseline model. The model is trained to always output scores, even if the data is well outside the training distribution. This is an existing issue with most NLP models and can be mitigated by additional checks for domain shift.

 \section*{Acknowledgments}
 This research is supported in part by the Office of the Director of National Intelligence (ODNI), Intelligence Advanced Research Projects Activity (IARPA), via the HIATUS Program contract \#2022-22072200006, the Defense Advanced Research Projects Agency with award HR00112220046, and NSF IIS 2048211. The views and conclusions contained herein are those of the authors. They should not be interpreted as necessarily representing the official policies, either expressed or implied, of ODNI, IARPA, or the U.S. Government. We would like to thank all the USC INK research lab collaborators for their constructive feedback on the work.

\bibliography{custom}
\bibliographystyle{acl_natbib}

\clearpage

\appendix
\section{Evaluation Datasets}
\label{app:dataset_statistics}
In this section, we describe the datasets used in our evaluation. We mention some important challenges that make these datasets useful for benchmarking the entailment verification task. Please refer to Table \ref{tab:dataset_statistics} for these datasets' train/dev/test statistics.

\ExecuteMetaData[sections/tables.tex]{dataStatistic}

\paragraph{Natural Language Inference datasets} NLI is an obvious choice of data source as it is a more general case of the entailment verification problem. While converting an NLI dataset for our task, we merge the \textit{neutral} and \textit{contradict} labels to the \textit{not support} label.\footnote{We note that merging the labels for the NLI datasets in this benchmark is valid, since the underlying meaning of the `neutral' label is `no sufficient evidence to support' in all three datasets. NLI datasets that don't follow this label definitions cannot be modified in this manner.} We use the following NLI datasets in our benchmark:
\begin{itemize}[itemsep=1pt,topsep=1pt]
	\item \textbf{\wanli{}} \cite{wanli}: This is a new NLI dataset built using worker and AI collaboration. This challenging dataset improves over the existing NLI dataset MultiNLI \cite{mnli}. The test split is used when doing the evaluation.
	\item \textbf{\fever{}} \cite{fever_nli}: This is a modification of the original FEVER dataset \cite{fever} in which the claim is paired with textual evidence from Wikipedia to convert it into an NLI format dataset. This pairing uses existing state-of-the-art evidence extraction systems to find relevant evidence for each claim. Premises in this dataset typically contain multiple sentences, which is one of our focus areas. As the test split is unavailable, we report results on the dev split for evaluation.
	\item \textbf{\anli{}} \cite{anli}: This is a large-scale NLI dataset that was collected using an adversarial human-and-model-in-the-loop procedure. Like \fever{}, this dataset tests factual knowledge, and the premises typically contain multiple sentences. During evaluation, the test split is considered.
\end{itemize}

\ExecuteMetaData[sections/figures.tex]{contextualQA}

\paragraph{Contextual QA datasets} Next, we consider QA datasets where the task is to answer a question based on a given context and some options. We use an off-the-shelf QA-to-statement converter model \cite{chen-etal-2021-nli-models} to generate a hypothesis statement for each question option pair. Then, the hypothesis corresponding to the correct choice is marked as ``\textit{support}'', while the rest are marked as ``\textit{not support}'' to create the entailment verification dataset. We depict this process in Figure \ref{fig:contextual_qa}. The green box is the valid hypothesis and the red ones are the invalid hypothesis corresponding to the given context. Overall, we include the following datasets from this category:
\begin{itemize}[itemsep=1pt,topsep=1pt]
	\item \textbf{\cosmosqalong{}} (\cosmosqa{}) \cite{cosmosqa}: This dataset contains multiple-choice questions (MCQs) that require an understanding of commonsense-based reading comprehension to answer a question. The key challenge in this dataset is understanding people’s everyday narratives described in the context that can have some missing commonsense knowledge that needs to be inferred implicitly. Since the test split is missing, we evaluate models on the dev split instead.
	\item \textbf{\siqalong{}} (\siqa{}) \cite{socialiqa}: Similar to \cosmosqa{}, this is another MCQ benchmark for commonsense reasoning about social situations that probes emotional and social intelligence in a variety of everyday situations. This dataset has more nuanced commonsense knowledge requirements, which makes it a challenging dataset for our task. Similarly, results on the dev split are reported, given that the test is missing.
	\item \textbf{\dream{}} \cite{dream}: This is a dialogue-based reading comprehension MCQ dataset that focuses on multi-turn dialogue understanding. Here, the unique challenge is inferring the events discussed across long, multi-turn dialogues. During evaluation, we use the test split as it is available.
	\item \textbf{\boolq{}} \cite{boolq}: This is a True/False QA dataset consisting of aggregated queries to the Google search engine. Questions in this dataset require complex and difficult entailment-like inference to solve, making it a good set for evaluation. The test split is lacking for this dataset, and we can only report results on the dev split.
	\item \textbf{\race{}} \cite{race}: The reading comprehension dataset from examinations (RACE) is one of the most popular machine reading comprehension datasets containing questions from English exams for middle and high school students. These questions are designed by domain experts for testing specific human reading skills, thus making it a good evaluation set for our task. We report evaluating results on test split for this dataset.
\end{itemize}

\paragraph{Rationale datasets} Lastly, we consider data sources where human-annotated explanations are available that justify the original hypothesis (or the correct option, in the case of QA datasets). In this case, we use the rationales as the premise. We use the following datasets:
\begin{itemize}[itemsep=1pt,topsep=1pt]
	\item \textbf{\entailer{}} \cite{entailer}: This dataset contains entailment-style statements and corresponding rationales obtained from EntailmentBank dataset \cite{entailmentbank} and crowdsourcing. The dataset mainly contains science domain statements and tests simple deductive reasoning skills whereby sentences from the premise must be combined to either support or refute the hypothesis. The test split is also missing for this dataset, and we can only evaluate it on the dev split.
	\item \textbf{\ecqa{}} \cite{ecqa}: This is a human-annotated explanation dataset for CommonsenseQA \cite{csqa}. We only use the explanations for the correct choice as the explanations for the incorrect choices are often trivial. It is a complete dataset and by convention, we use the test split for evaluation.
\end{itemize}

\section{Details on LLM Evaluation}
\label{app:llm_details}
We evaluate two types of LLMs on the task of entailment verification, as categorized below:

\subsection{Task-Finetuned models}
\label{app:task_finetuned}
In this category, the models considered are already finetuned for either NLI or the exact entailment verification task itself. We evaluate two models in this category.
\paragraph{\roberta{}}\cite{anli, roberta} 
This is a strong pre-trained RoBERTa-Large model with the corresponding model card on HuggingFace\footnote{\url{https://huggingface.co/models}} \cite{wolf2020huggingfaces} called ``ynie/roberta-large-snli\_mnli\_fever\_anli\_R1\_R2\_R3-nli''. It is a specifically pre-trained RoBERTa-Large for NLI task and includes the combination of SNLI \cite{snli}, MNLI \cite{mnli}, FEVER \cite{fever_nli} and ANLI \cite{anli} datasets as the training data. Hence, this incurs a potential data leakage problem, as we also test it on FEVER and ANLI. To some extent, it explains the strong performance of \roberta{} on NLI datasets in Table \ref{tab:human_results}.

The model is used as a classifier in evaluation, and three classes are available. Class 0 corresponds to ``\textit{entail}'', class 1 corresponds to ``\textit{neutral}'' and class 2 means ``\textit{not entail}''. In our experiment setting, we only regard class 0 as the ``\textit{Yes}'' label and combine the remaining two classes to be the ``\textit{No}'' label.

\paragraph{\entailerm{}}\cite{entailer}
Entailer is a T5-based model \cite{raffel2019exploring} trained to answer hypotheses by building proof trees containing chains of reasoning. It can either generate valid premises for a given hypothesis or predict a score for a given premise and hypothesis. We evaluate \entailermx{} (model name ``allenai/entailer-11b'' on HuggingFace) in our experiments. Similarly, \entailer{} dataset \cite{entailmentbank} is in the training set, making this model very competitive when evaluating the same dataset. 

We strictly follow the official implementation of the model to acquire class labels.\footnote{\url{https://github.com/allenai/entailment_bank/blob/main/entailer.md}} The ``entailment\_verifier'' is called to decide if the hypothesis can be implied from the premise. If the answer is ``\textit{True}'', then the class label will be ``\textit{Yes}'' and vice versa.

\ExecuteMetaData[sections/tables.tex]{promptSensitivity}

\ExecuteMetaData[sections/tables.tex]{fewshotResults}

\subsection{Instruction-Finetuned models}
\label{app:instruction_finetuned}
These are the more recent ``general-purpose'' language models trained on a collection of NLP tasks described using instructions, leading to generalization abilities to solve unseen tasks described using new instructions. The models included in our evaluation from this category are described below.
\paragraph{\flantxxl{}}\cite{flant5}
It is instruction-tuned from T5 \cite{raffel2019exploring} on 1.8K+ tasks. We adopt a publicly available version on HuggingFace \cite{wolf-etal-2020-transformers} with model card name ``google/flan-t5-xxl''. \flantxxl{} is also exposed to data leakage issues. \boolq{} \cite{boolq}, \ecqa{} \cite{ecqa}, and \anli{} \cite{anli} have appeared in its training data. However, this is not a serious problem in the finetuning stage because we transform original datasets into entailment verification format before using them for model finetuning.

We extract labels from the model by focusing on the output probabilities of two words, ``\textit{Yes}'' and ``\textit{No}''. After applying the softmax function to those two probabilities, we finalize the label as the word with a probability larger than a given threshold, as described in Equation \ref{eq:score}.

\paragraph{\gptthree{}}\cite{gpt3}
It is a general-purpose autoregressive decoder-only LLM accessible via the OpenAI Completions API.\footnote{\url{https://platform.openai.com/docs/api-reference/completions}} We utilize ``text-davinci-003'' in OpenAI's API for evaluation, and the label determination procedure is similar to the one in \flantxxl{}.

\paragraph{\gptfour{}}\cite{gpt4}
It is the latest generative model published by OpenAI, which is optimized for creativity and long context inputs. It is accessible via the OpenAI Chat API.\footnote{\url{https://platform.openai.com/docs/api-reference/chat}} We adopt plain ``gpt-4'' in OpenAI's API for our experiments. Unlike other models, the output probabilities are not accessible. Thus, we constrain the model to predict a single token as its generated prediction. In the ideal case, we can directly use the output text as the label if ``\textit{Yes}'' or ``\textit{No}'' is produced. If some other token is generated, we choose one of the labels at random. We note that this occurrence is rare in our \gptfour{} evaluation runs.

\subsubsection{Prompt Robustness Evaluation}
To assess the robustness of the model in section \ref{sec:baselines}, we design different prompt formats but hold the order of premise and hypothesis in the prompt unchanged. Table \ref{tab:prompt_sensitivity} presents all prompts we tested with \flantxxl{} and their corresponding averaged results across datasets. The results suggest that the \flantxxl{} model is robust to the variation in the prompt and yields relatively consistent results. This characteristic is maintained in \ourclass{} and \ourrank{} as well since they are developed over based on \flantxxl{}.

\subsubsection{Few-Shot Evaluation}
Few-shot is an effective and promising strategy when testing the performance of a model \cite{gpt3}. We also include this analysis by randomly picking two examples from \entailer{} as demonstrations and incorporating them into the prompt. We test \flantxxl{}, \ourrank{}, and \gptfour{} in this setting and represent results in Table \ref{tab:fewshot_results}. The few-shot setting yields promising improvement for \flantxxl{}, substantiating that the few-shot is a beneficial approach to teaching the prompt to the model. However, it is not as helpful as our finetuning strategies, which give even better performance. On the other hand, few-shot does not bring significant gains for \ourrank{}, suggesting that finetuning has already helped the model have a comprehensive understanding of the prompt, and extra demonstrations are unnecessary. As for the \gptfour{}, simply using examples from \entailer{} and applying the same prompt for all datasets seem detrimental.

\ExecuteMetaData[sections/tables.tex]{majorityResults}

\section{Majority Prediction  and Label Imbalance}
\label{app:majority_baseline}
In Table \ref{tab:majority_results}, we show the performance of an oracle model that predicts the most frequent label in a dataset. For a label-balanced dataset, the macro-F1 score of such a majority prediction model would be 0.67 (precision 0.5 and recall 1.0). The datasets in the evaluation set have some label imbalance, as evidenced by the lower majority label prediction scores. Since we convert existing 3-class NLI and multi-choice QA datasets into our binary classification task format, it inherently has more \textit{not support} labels. We have more \textit{support} instances for the rationale datasets since the dataset creators usually only annotate the rationale for the right choice. Specifically, the \ecqa{} dataset only has positive instances, leading to a 1.0 macro-F1 score for majority prediction (\textit{support} label). Since it has all \textit{support} labels, any model predicting even a single \textit{non support} label gets penalized severely, as is seen in \ecqa{} results. Because of this label imbalance in the datasets, we report the macro-F1 scores instead of accuracy or micro-F1 metrics.

\section{Debatable Cases in Contextual QA}
\label{app:debatable_example}
We include two examples of \siqa{} dataset in Table \ref{tab:contextual_examples} to illustrate the debatable hypotheses in Contextual QA datasets. These examples demonstrate that some less probable hypotheses can also be considered plausible in certain situations. In the first example, option ``run screaming'' best describes the audience's reactions after listening to a scary ghost story. However, there are also situations where people are so terrified and start to cry, making the option ``cry'' somehow debatable. Similarly, in the second example, it is common sense that people will express gratefulness after someone else builds a house for them. But it is reasonable to say that Quinn is intrigued by the house after seeing the amazing architecture. In both cases, apart from the most adequate one, there are still options leading to some partially supported hypotheses, which explains the low human annotation agreement in these cases.

\ExecuteMetaData[sections/tables.tex]{contextualExamples}

\ExecuteMetaData[sections/figures.tex]{reasoningAnalysis1}
\section{Human Evaluation and Analysis}
\label{app:human_ann}
We adopted Amazon Mechanical Turk (MTurk) \cite{amt} for data collection. Two annotation formats (Figure \ref{fig:eval_annotate} and Figure \ref{fig:reason_annotate}) were devised for the human evaluation task and reasoning type analysis task, respectively. During the annotations, each annotator was compensated according to \$15/hour per the U.S. minimum wage.

\subsection{Human Evaluation Details}
\label{app:human_ann_eval}
In each HIT, the annotator was presented with a format exactly like Figure \ref{fig:eval_annotate}, including detailed task descriptions and label explanations. Annotators were expected to read the premise and claim first, then determine the supportiveness of the claim based on the premise and choose the corresponding label. Initially, we only provided three labels --- ``\textit{support}'', ``\textit{irrelevant}'' and ``\textit{contradict}''. But later, we realized that annotators could not explicitly identify labels for some ambiguous instances where the premise only partially supported or contradicted the claim. Hence, we introduced two weak labels (``\textit{partially support}'' and ``\textit{partially contradict}'') to remedy this issue. When collating results for analysis, we internally combined ``\textit{support}'' and ``\textit{partially support}'' to be ``\textit{support}'', and the rest to be ``\textit{not support}'', aligning to the standard entailment verification (EV) setup. Each instance was annotated by 3 MTurk annotators, and a majority verdict determined the label. To ensure annotation quality, we conducted two rounds of qualification for annotators using the finalized template. In the first round, we used ten questions from the datasets and evaluated 400 mTurk annotators. We retained 100 annotators from this batch with an accuracy greater than 70\%. Among remained ones, we repeated this process for another ten questions and selected annotators with an accuracy greater than 80\%. Finally, we retained 35 annotators who had annotated the human evaluation set. And the Fleiss’s kappa score \cite{fleissscore} we got was 0.6, indicating a moderate level of agreement among annotators. 

\ExecuteMetaData[sections/figures.tex]{evalAnnotate}

\subsection{Reasoning Type Annotation Details}
\label{app:human_ann_reasoning}
We annotate the reasoning type of the 100 sampled instances for each dataset with absolute macro-F1 difference $> 0.1$ in Table \ref{tab:human_results}. We make this choice intending to attribute the largely misaligned datasets (namely, \anli{}, \cosmosqa{}, \siqa{}, and \entailer{}) since some random noise in the annotation and sampling process can potentially also cause some misalignment.

In every HIT, we used Figure \ref{fig:reason_annotate} as the reasoning type annotation format. The instructions and label explanations were explicitly stated at the beginning of the HIT. For task 1, after reading the premise and the claim, annotators had to decide whether the supportiveness of the claim could be decided by only referring to the information in the premise. If yes, annotators needed to choose the corresponding difficulty level of reasoning about the supportiveness of the claim in task 2. Otherwise, the type of missing information had to be decided in task 1, and task 2 did not apply to these cases. We aggregated answers from the two tasks and categorized them into four types. Both \textbf{R1} and \textbf{R2} were types where the premise contained all necessary information. The difference was that the difficulty level of reasoning for \textbf{R1} was \textit{Easy} while for \textbf{R2} was \textit{Moderate}. We combined \textit{Missing Entity-grounded Information} and \textit{Missing Commonsense Information} to be \textbf{R3}. Finally, the \textit{Missing Localized Information} label corresponded to \textbf{R4}.

Given that some strong NLP background knowledge was required to understand task descriptions and label explanations, we recruited Computer Science graduate students instead of the general mTurk workers to finish this annotation. Every instance was assigned to 2 students, and the majority vote determined the label. The Fleiss’s kappa score for this job was 0.62, showing a substantial inter-annotator agreement.

\ExecuteMetaData[sections/figures.tex]{reasonAnnotate}
\subsection{Reasoning Type Examples}
\label{app:reasoning_examples}
Table \ref{tab:reasoning_examples} incorporates three examples from each reasoning type, providing more insight into those types.

In the third example of \textbf{R1}, the first sentence in the premise states that ``\textit{iron oxide}'' comes from ``\textit{oxygen}'' and ``\textit{rust}''. The second sentence shows those two substances are ``\textit{gases}'' at room temperature. Therefore, combining them will be sufficient to entail the hypothesis. 

However, unlike \textbf{R1}, the third example of \textbf{R2} requires three steps of reasoning that ``De Baandert was a multi-use stadium.'', ``It was mostly used for football matches.'', and ``The stadium was able to hold 22,000 people.''. The hypothesis can be disproved with those steps because ``22,000 people'' is just the maximum capacity.

As for the first example of \textbf{R3}, some missing commonsense information like ``\textit{it is not wise to give more money to a person who keeps playing in a detrimental situation.}'' should be combined with the premise to disprove the hypothesis.

In the first example of \textbf{R4}, the next movement of ``\textit{Addison}'' is the missing information specific to the context depicted by the premise. The hypothesis can not be directly disproved without that piece of information.

\subsection{Reasoning Type Analysis}
\label{app:reasoning_analysis}
We depict the aggregated results of the reasoning type annotation for each dataset in Figure \ref{fig:reasoning_analysis1}. Here, the first row shows the frequency of each reasoning type in that dataset, and the corresponding plot in the second row compares the human and \gptfour{} macro-F1 scores. We fade out the columns with data percentages less than 5\% because such low frequency might not lead to conclusive observations.

We note that type \textbf{R1} is most prominent in \entailer{} dataset. Here, we observe that humans are significantly better than models. This shows that humans are usually more consistent with simple deductive reasoning. Similar findings about consistency in human deductive reasoning skills have been reported in prior works \cite{robustlr,nguyen2023negation}.

Reasoning type \textbf{R2}, which requires more complex reasoning, is dominant in \anli{} and \cosmosqa{}. For this type, we find that models are superior to humans. Complex reasoning requires two skills: understanding multiple relevant information in the premise and then using them for reasoning. We hypothesize that models are stronger context processors than humans because they have been trained on long-context data \cite{gpt4}.

The reasoning type \textbf{R3} is present in \anli{}, \cosmosqa{}, and \siqa{}. From Table \ref{tab:dataset_comparisons}, we know that \anli{} mostly require entity-grounded knowledge, whereas \cosmosqa{} and \siqa{} specifically test commonsense knowledge. Here, we find that humans are stronger than models in commonsense knowledge (\cosmosqa{} and \siqa{}), whereas models are better in \anli{} that requires entity-grounded knowledge. This shows that humans can infer missing social/commonsense knowledge more easily since these are inherently known to humans. In contrast, models can retrieve the entity-grounded knowledge stored in their parameters more efficiently.

Lastly, we find that the reasoning type \textbf{R4}, indicating missing localized knowledge, is also prominent in \anli{}, \cosmosqa{}, and \siqa{}. Here, we find that the trends are a bit mixed. We find that for \siqa{}, humans are better at recognizing missing localized knowledge, but in \cosmosqa{}, models outperform humans. This is likely because \siqa{} typically contains short contexts based on everyday social situations that are easier for humans to understand. In contrast, \cosmosqa{} has longer contexts with rarer situations requiring more complex understanding.

Overall, we conclude that \gptfour{} outperforms humans in complex deductive reasoning and situations involving entity-grounded knowledge, whereas humans are more consistent at simple reasoning and situations requiring commonsense knowledge.

\section{Finetuning LLMs}
\label{app:finetuning_llms}
In this section, we describe more details about the training dataset used for finetuning, our negative data collection strategy that was used in the ranking formulation, and other finetuning details.

\subsection{Training Dataset Selection}
\label{app:training_data_selection}
To train the \flantxxl{} model, we create a training dataset using representative datasets from each category. We pick the \anli{}, \race{}, and \ecqa{} datasets to represent NLI, contextual QA, and rationale categories, respectively. We select the datasets with diverse entailment challenges and aim to maximize the total training data. We note that the amount of training data is quite low for the rationale category. Thus, we also include the StrategyQA \cite{strategyqa} dataset in the training set to alleviate this. Similar to \boolq{}, StrategyQA has a Yes/No type of questions and their corresponding explanations. We convert the question-answer pair into a hypothesis using the QA-to-statement converter \cite{chen-etal-2021-nli-models} as described in Section \ref{sec:datasets}.

\begin{BOX}
	\centering
	\noindent\fbox{%
		\parbox{0.95\columnwidth}{%
			For a given premise and a valid hypothesis, generate five alternate hypotheses contradicted by the premise. Try to avoid using the negation words such as ``not'', ``never'', etc. The output should be numbered from 1 to 5.\\
			\textbf{Premise}: \{\textit{premise}\}\\
			\textbf{Hypothesis}: \{\textit{hypothesis\}}
		}%
	}
	\caption{\label{box:neg_prompt} Prompt format for generating alternate negative hypothesis for a given premise-hypothesis pair. Please refer to Section \ref{app:ranking} for details.}
\end{BOX}

\ExecuteMetaData[sections/tables.tex]{cotResults}

\subsection{Negative Data Collection for Ranking}
\label{app:ranking}
In the ranking formulation, for a given premise and hypothesis pair $(p, h)$, we need to find some weaker hypothesis $h^\prime$ to use the ranking loss defined by Equation \ref{eq:ranking_loss}. We collect such weaker hypotheses in two ways and then combine them to form the training data. The two techniques are described below:
\begin{itemize}[itemsep=1pt,topsep=1pt]
	\item \textbf{Using incorrect options}: The contextual QA category has naturally occurring negative data. For a given question and choices, we pair the hypothesis corresponding to the correct option with all other hypotheses corresponding to the wrong options to create the ranked data.
	\item \textbf{\gptthree{} prompting}: The other way we generate negative data is by prompting \gptthree{}. Specifically, we use the prompt format shown in Box \ref{box:neg_prompt} to generate alternate hypotheses contradicted by the original premise. We only select premise and hypothesis pairs that originally have \textit{support} label. \gptthree{} generated hypotheses are then considered negative samples and paired with the original hypothesis. We repeat this for all the training datasets (\anli{}, \race{}, \ecqa{}, and StrategyQA).
\end{itemize}

\subsection{Hyperparameters and other details}
\label{app:hyperparams}
During training, we select the learning rate from the set $\{7e^{- 5}, 1e^{- 4}, 2e^{- 4}\}$, per GPU batch size from the set $\{6, 8\}$, margin $m$ in Equation \ref{eq:ranking_loss} from the set $\{0.2, 0.3, 0.5\}$, and warmup ratio $0.1$. The model is trained for 1400 steps on a cluster of 8 A6000 GPUs. We evaluate the model every 200 steps and save the checkpoint if the model shows improvements on a held-out development set.

\section{Chain-of-Thought Filtering}
\label{app:cot_filtering}
We study three variants of CoT Filtering as mentioned below:
\begin{itemize}[itemsep=1pt,topsep=1pt]
	\item $\mathcal{B} + \textrm{SC}$: This is the self-consistency baseline. Here, $\mathcal{B}$ is the base model used to sample CoTs. We sample 40 CoTs for each instance before computing the majority predicted label.
	\item $\mathcal{B} + \textrm{\flantxxl{} + SC}$: In this, we use a pre-trained \flantxxl{} for filtering out the inconsistent rationales before the majority voting. We keep the top-5 rationales after scoring them using \flantxxl{}.
	\item $\mathcal{B} + \textrm{\ourrank{} + SC}$: This is the same as above, but instead, we use our ranking-finetuned \flantxxl{} model for filtering.
\end{itemize}
Following \cite{wang2023selfconsistency}, we use four different base CoT model: \ultwo{} \cite{ul2}, \codexone{} \cite{gpt3}, \lamda{} \cite{lamda}, and \chatgpt{} \cite{chatgpt}. Further, we compute the CoTs and analyze the performance of the above methods for three multi-choice QA datasets, namely, CommonsenseQA \cite{csqa} and AI2 Reasoning Challenge \cite{arc} (easy (\arce{}) and challenge (\arcc{}) variants). Please refer to \cite{wang2023selfconsistency,wei2022chain} and the associated code\footnote{\url{https://openreview.net/attachment?id=1PL1NIMMrw&name=supplementary_material}} for details on the CoT prompt formats. The results are shown in Table \ref{tab:cot_results}. We note a consistent improvement between the three variants, with \ourrank{} model performing the best. This demonstrates the advantage of our entailment finetuning approach. Please refer to Section \ref{sec:filtering_cot} for more findings.

\ExecuteMetaData[sections/figures.tex]{k-selection}

\subsection{Ablation of Top-k Filtering}
\label{app:top-k_analysis}
We examine our top-$k$ selecting strategy by choosing different values of $k$ and compare their results. We pick up $k$ from $\{3, 5, 10, 20, 30\}$ and present the corresponding aggregated results in Figure \ref{fig:k-selection}. The general trend is that, as $k$ increases, the model performance increases first and then starts to decrease, reaching the peak at $k=5$. When $k$ is small, there are only a limited number of CoT rationals, and the majority voting process is vulnerable to potential outliers. On the other hand, if k is too large, then many noisy results are included, leading to poorer performance.

\subsection{Examples of Filtered CoTs}
\label{app:cot_examples}
Table \ref{tab:filter_examples} presents three CoT reasoning examples, each including two outputs that are kept and three that are filtered out by ranking. According to the table, outputs supported by strong rationales are ranked highly and kept. On the other hand, if the rationale is irrelevant to the prediction (like rationale 3 in example 1), the rationale itself is incomplete (like rationale 5 in example 1), or the rationale supports another option rather than the prediction (like rationale 4 in the example 1), then such output has a low entailment score leading to a lower ranking and getting filtered out.

\ExecuteMetaData[sections/tables.tex]{reasoningExamples}

\ExecuteMetaData[sections/tables.tex]{filterExamples}

\end{document}